%% file: conference_101719.tex
\documentclass[conference,a4paper]{IEEEtran}
\IEEEoverridecommandlockouts

\input{commands}

\usepackage{cite}
\usepackage{amsmath,amssymb,amsfonts}
\usepackage{algorithmic}
\usepackage{graphicx}
\usepackage{textcomp}
\usepackage{xcolor}
\usepackage{booktabs}
\def\BibTeX{{\rm B\kern-.05em{\sc i\kern-.025em b}\kern-.08em
    T\kern-.1667em\lower.7ex\hbox{E}\kern-.125emX}}
\begin{document}

\title{Multiscale Vision Transformer With\\ Deep Clustering-Guided Refinement for\\ Weakly Supervised Object Localization\\
\thanks{This work was supported in part by the National Research Foundation of Korea (NRF) grant funded by the Korea government(MSIT) (No.RS-2023-00252434). Corresponding author: Byeongkeun Kang (byeongkeun.kang@seoultech.ac.kr).}
}

\author{\IEEEauthorblockN{David Minkwan Kim}
\IEEEauthorblockA{Department of Computer Science \\
Hanyang University \\
Seoul, Republic of Korea \\
}
\and
\IEEEauthorblockN{Sinhae Cha \hspace{0.85cm} Byeongkeun Kang}
\IEEEauthorblockA{Department of Electronic Engineering \\
Seoul National University of Science and Technology \\
Seoul, Republic of Korea \\
}
}

\maketitle

\begin{abstract}
This work addresses the task of weakly-supervised object localization. The goal is to learn object localization using only image-level class labels, which are much easier to obtain compared to bounding box annotations. This task is important because it reduces the need for labor-intensive ground-truth annotations. However, methods for object localization trained using weak supervision often suffer from limited accuracy in localization. To address this challenge and enhance localization accuracy, we propose a multiscale object localization transformer (MOLT). It comprises multiple object localization transformers that extract patch embeddings across various scales. Moreover, we introduce a deep clustering-guided refinement method that further enhances localization accuracy by utilizing separately extracted image segments. These segments are obtained by clustering pixels using convolutional neural networks. Finally, we demonstrate the effectiveness of our proposed method by conducting experiments on the publicly available ILSVRC-2012 dataset.
\end{abstract}

\begin{IEEEkeywords}
weakly-supervised object localization, weakly-supervised learning, vision transformer, neural networks
\end{IEEEkeywords}

\input{sections/1_introduction}

\input{sections/3_proposed_method}

\input{sections/4_results}

\section{Conclusion}
We introduced a novel multiscale object localization transformer for accurate object localization in input images using patch embeddings across various granularities. We then proposed the deep clustering-guided refinement method, refining pixel-level activation values using image segments. These segments were generated via the deep pixel clustering method, leveraging convolutional neural networks and cluster associations. We evaluated our approach's effectiveness by measuring Top-1, Top-5, and GT-known localization accuracies on the ILSVRC-2012 dataset, affirming its efficacy.

\bibliographystyle{IEEEtran}
\bibliography{mybibfile}

\end{document}

%% file: commands.tex
\newcommand{\etal}{{\it {et al}.} }

\usepackage{subfigure}
\usepackage{amsmath}

\usepackage{amssymb}
\usepackage{xcolor}
\usepackage{tabularx} 
\usepackage{multirow}
\usepackage{tabu}
\usepackage{pbox}
\usepackage{amssymb}
\usepackage{amsfonts}
\usepackage{bbm}
\usepackage{graphicx}
{\begin{list}               
    {$\bullet$ \hfill}{
        \setlength{\leftmargin}{\parindent}
        \setlength{\parsep}{0.04\baselineskip}
        \setlength{\itemsep}{0.5\parsep}
        \setlength{\labelwidth}{\leftmargin}
        \setlength{\labelsep}{0em}}
    }
{\end{list}}

\providecommand{\cref}[1]{Chapter~\ref{#1}}

\providecommand{\fref}[1]{Fig.~\ref{#1}}
\providecommand{\tref}[1]{Table~\ref{#1}}

\providecommand{\R}{\ensuremath{\mathbb{R}}}

\renewcommand{\vec}[1]{\ensuremath{\boldsymbol{#1}}}
\providecommand{\mat}[1]{\ensuremath{\boldsymbol{#1}}}

\providecommand{\calL}{\mathcal{L}}

\providecommand{\mA}{\mat{A}}

\providecommand{\mI}{\mat{I}}

\providecommand{\mM}{\mat{M}}

\providecommand{\mS}{\mat{S}}

\providecommand{\vk}{\vec{k}}

\providecommand{\vp}{\vec{p}}
\providecommand{\vq}{\vec{q}}

\providecommand{\vs}{\vec{s}}

\providecommand{\vv}{\vec{v}}

\providecommand{\vy}{\vec{y}}
\providecommand{\vz}{\vec{z}}



%% file: sections/1_introduction.tex
\section{INTRODUCTION} \label{sec:introduction}
Recently, there has been considerable interest in weakly-supervised object localization (WSOL) techniques due to their ability to effectively reduce the need for labor-intensive human annotations in training neural networks for object localization~\cite{Oquab2015, cam2016, Gao2021, fdcnet}. Unlike supervised learning-based methods that depend on training images, image-level class labels, and bounding boxes, WSOL methods solely utilize images and image-level class labels~\cite{Gao2021}. As annotating bounding boxes requires substantially more effort than annotating image-level class labels, WSOL methods offer a promising solution to minimize human annotation efforts. Therefore, we focus on investigating WSOL methods.

Oquab \etal introduced one of the earliest approaches for WSOL in~\cite{Oquab2015}, which employed convolutional neural networks (CNNs) along with a sliding window and max-pooling. To avoid the sliding window for improved efficiency, Zhou \etal used global average pooling instead of max-pooling and generated a dense localization map which is referred to as the class activation map (CAM)~\cite{cam2016}. However, due to the lack of supervision for localization during training, these approaches tend to localize only the most discriminative part of an object and struggle with accurately identifying the entire object region. Therefore, in the subsequent effort, researchers have explored various techniques to localize the entire object region rather than just the most discriminative part. These methods include using pseudo labels~\cite{Zhang2020}, adversarial erasing~\cite{ACoL2018}, multiple feature maps~\cite{Xue2019}, and alternative architectures~\cite{Gao2021}. 

Considering alternative architectures, Gao \etal introduced a vision transformer~\cite{visionTransformer2020}-based method to leverage self-attention mechanisms for long-range dependencies rather than local receptive fields in convolutional layers~\cite{Gao2021}. 


\input{sections/fig_framework}

In this paper, we also investigate a vision transformer-based framework that takes advantage of self-attention mechanisms to capture long-range dependencies for localizing an object's entire region. Given that previous methods still have limitations in accurately localizing the entire object region, we propose a multiscale object localization transformer (MOLT) that extracts patch embeddings using various receptive fields. Then, by aggregating the output patch embeddings from the transformers, the multiscale transformer effectively localizes object components, ranging from coarse discriminative regions to fine details. Additionally, we introduce a deep clustering-guided refinement method to enhance object localization even further. This method involves refining pixel-level activation values using the activation values of the corresponding image segment. These segments are obtained by clustering pixels using convolutional neural networks.

In summary, we propose a WSOL framework that consists of a multiscale object localization transformer, deep pixel clustering, and a deep clustering-guided refinement method, as illustrated in~\fref{fig:framework_inference}. The proposed multiscale transformer comprises three object localization transformers and is to localize an object's entire location rather than only the most discriminative part by activating the object in varying granularities. The deep clustering-guided refinement is to further improve localization accuracy using separately extracted image segments. Finally, we verify the effectiveness of the proposed method by measuring Top-1, Top-5, and GT-known localization accuracies using the ILSVRC-2012 dataset~\cite{ILSVRC15}.

%% file: sections/fig_framework.tex
\begin{figure*}[!t] 
\begin{center}
\begin{minipage}{0.49\linewidth}
\centerline{\includegraphics[scale=0.65]{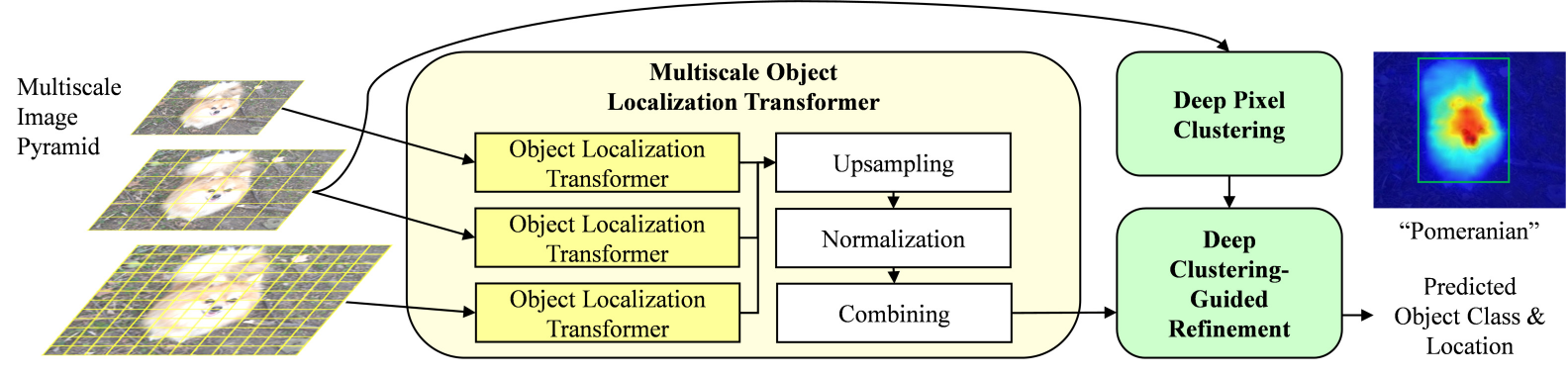}}
\end{minipage}
   \caption{Illustration of the proposed framework. The proposed framework consists of a multiscale object localization transformer, deep pixel clustering, and deep clustering-guided refinement. The multiscale transformer comprises three object localization transformers. The refinement method takes the combined class activation map from the multiscale transformer and image segments from deep pixel clustering, and predicts the object class and location.}
\label{fig:framework_inference}
\end{center}
\end{figure*}

%% file: sections/3_proposed_method.tex
\section{PROPOSED METHOD}
\label{sec:method}
We propose a multiscale object localization transformer that predicts both the object's class and location in an input image. Furthermore, we introduce a deep clustering-guided refinement method that enhances localization accuracy by refining activation values at the pixel level.

\subsection{Multiscale Object Localization Transformer}
\label{sec:wsol_network}
The multiscale object localization transformer (MOLT) employs a vision transformer architecture~\cite{visionTransformer2020} to predict both the object class and its location given an input image. To accurately predict them, we first construct a multiscale image pyramid by resizing the image into three different resolutions ($H_1 \times H_1$, $H_2 \times H_2$, $H_3 \times H_3$), as shown in~\fref{fig:framework_inference}. This is to explicitly extract embeddings at varying granularities. Subsequently, each image in the pyramid is fed into the corresponding object localization transformer, which is based on the architecture presented in~\cite{Gao2021}.

Given the $i$-th image $\mI_i$ in the pyramid, it is first cropped into $N_i \times N_i$ patches, where each patch has a size of $\hat{H}_i \times \hat{H}_i$. These patches are then linearly projected using learnable parameters to obtain the patch embedding $\hat{\vz}_i^0 \in \R^{{N_i}^2 \times D}$. The patch embedding is then appended after a class token which is a $D$-dimensional vector, resulting in the token embedding $\tilde{\vz}_i^0 \in \R^{L_i \times D}$, where $L_i={N_i}^2+1$. Finally, the input token embedding $\vz_i^0 \in \R^{L_i \times D}$ is obtained by adding $\tilde{\vz}_i^0$ with the learnable position embedding and by applying layer normalization~\cite{ba2016layer}.

The input token embedding $\vz_i^0$ is then processed by the transformer encoder, which comprises consecutive transformer blocks. In the $(b+1)$-th block, the input embedding $\vz_i^b$ undergoes a sequence of processing steps: layer normalization, multi-head self-attention (MSA) with a residual connection, and a multi-layer perceptron (MLP) with a residual connection. The MLP consists of two layers and employs the Gaussian error linear units (GELU) function~\cite{hendrycks2023gaussian} for intermediate non-linearity. The output $\vz_i^{b+1} \in \R^{L_i \times D}$ from the MLP serves as the input embedding for the subsequent block.

The multi-head self-attention (MSA) consists of $M$ attention heads and employs query, key, and value-based self-attention. The normalized input sequence $\bar{\vz}_i^b$ is first divided into $M$ vectors, each with dimensions of ${L_i \times \frac{D}{K}}$. Then, for the $m$-th head, query $\vq_{i,m}^b$, key $\vk_{i,m}^b$, and value $\vv_{i,m}^b$ vectors are obtained by linearly projecting the $m$-th vector from $\bar{\vz}_i^b$ using learnable parameters. An attention matrix $\mA_{i,m}^b \in \R^{L_i \times L_i}$ is then calculated by multiplying the query vector $\vq_{i,m}^b$ with the transposed key vector $\vk_{i,m}^b$, by scaling the result using a predetermined scalar, and by applying a softmax function. The output vector for the head is computed by multiplying the attention matrix $\mA_{i,m}$ with the value vector $\vv_{i,m}^b$. The output of the MSA is obtained by concatenating the outputs from the self-attention heads and by applying a fully connected layer. As mentioned earlier, we employ the MSA with a residual connection, so the output is added to the input sequence $\vz_i^b$.

The output embedding sequence from the transformer encoder is reshaped into a tensor with dimensions of $N_i \times N_i \times D$, after excluding the output class token. Subsequently, the class score map $\mS_i \in \R^{N_i \times N_i \times C}$ is obtained by applying a 2D convolution layer to the reshaped tensor, where $C$ denotes the number of classes. Finally, the class score $\vs_i$ is computed by applying global average pooling to the class score map $\mS_i$. During inference, the argmax function is applied to the class score $\vs_i$ to predict the object class label.

To predict an object's location, a class activation map (CAM) is estimated using the attention maps $\mA_{i,m}^b$ and the class score map $\mS_i$. Firstly, the attention maps $\mA_{i,m}^b$ are averaged over the multiple heads ($m$) and summed over the transformer blocks ($b$), resulting in an averaged attention map $\bar{\mA}_{i} \in \R^{L_i \times L_i}$. Then, the foreground map $\mM_i^{fg}$ is estimated using the attention weights connecting from the $N_i \times N_i$ patch tokens to the class token. Finally, the CAM $\mM_i^{cam}$ is obtained by multiplying the foreground map $\mM_i^{fg}$ with the class score map $\mS_i$.

The CAMs $\mM_i^{cam}$ from the object localization transformers are then upsampled to match the highest resolution among them. Finally, each of the maps is normalized using min-max normalization, and the normalized maps are used to generate the combined CAM $\mM^{ccam}$. During inference, an object's location is predicted by identifying the bounding box of the region with activation values exceeding a certain threshold in $\mM^{ccam}$.

\noindent \textbf{Training.}
Because only image-level class labels are available, the MOLT is trained using these labels to predict an object's class and its location. While a typical image classification transformer predicts an object class using the output class token, this object localization transformer is designed to estimate the object class using the output embedding sequence. It is to explicitly depend on the entire patch embeddings rather than only the class token. Specifically, the probability $\vp$ of containing a certain object is estimated by applying a softmax function to the class score $\vs$, which was introduced earlier.

Then, the loss $\calL$ is computed using a cross-entropy loss function, which compares the predicted probability vector $\vp$ to the ground-truth class label $\vy$.
\begin{equation}
	\calL = - \sum_{c=1}^{C} \vy_c  \ln({ \sigma(\text{GAP}( \mS ))_{c}}) = - \sum_{c=1}^{C} \vy_c  \ln{\vp_{c}}
\label{eq:loss_class}
\end{equation}
where $C$ represents the total number of classes; GAP and $\sigma(\cdot)$ denote global average pooling and a softmax function, respectively.

\input{sections/fig_analysis_multiscale}

\noindent \textbf{Analysis.}
\fref{fig:analysis_multiscale} illustrates the class activation maps obtained from the MOLT using the ILSVRC-2012 dataset~\cite{ILSVRC15}. This visualization highlights the complementary nature of CAMs across the transformers with varying patch scales. CAMs generated by the transformer with low resolution emphasize the overall object location, while CAMs produced by the transformers with intermediate or high resolutions have an advantage in localizing the finer details of objects.

\subsection{Deep Clustering-Guided Refinement}
\label{sec:refinement}
Since CAMs tend to localize only the most discriminative region, we propose a deep clustering-guided refinement method in addition to the MOLT. While the combined CAM $\mM^{ccam}$ from the proposed transformer localizes objects more accurately than a single-scale transformer, as illustrated in~\fref{fig:analysis_multiscale}, this clustering-guided refinement can lead to even further improvements in localization accuracy.

\begin{figure}[!t] 
\begin{center}
\begin{minipage}{0.23\linewidth}
\centerline{\includegraphics[scale=0.25]{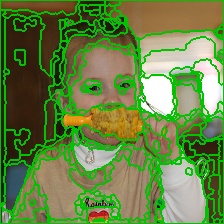}}
\end{minipage}
\begin{minipage}{0.23\linewidth}
\centerline{\includegraphics[scale=0.25]{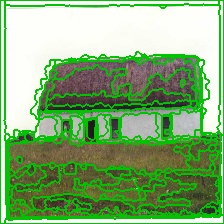}}
\end{minipage}
\begin{minipage}{0.23\linewidth}
\centerline{\includegraphics[scale=0.25]{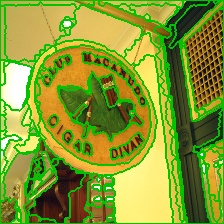}}
\end{minipage}
\begin{minipage}{0.23\linewidth}
\centerline{\includegraphics[scale=0.25]{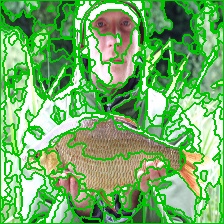}}
\end{minipage}
\caption{Visualization of deep pixel clustering results utilized in the refinement process. Green lines denote boundaries between regions.}
\label{fig:result_DIC}
\end{center}
\end{figure}

To refine the localization, we first cluster the pixels within an image into multiple regions using a neural network-based unsupervised clustering method, as illustrated in~\fref{fig:result_DIC}. Subsequently, we calculate the mean of the activation values of the pixels belonging to each cluster. Finally, the refined activation map is obtained through a weighted summation of $\mM^{ccam}$ and the mean activation value of the corresponding cluster for each pixel.

In more detail, we employ the deep image clustering method~\cite{dic2020Zhou}, which consists of a feature extractor and a deep clustering algorithm, for unsupervised pixel clustering. Due to the absence of ground-truth labels, the feature extractor is trained using a superpixel-based loss. We select the simple linear iterative clustering (SLIC) method~\cite{slic2012} to extract superpixels. The results of deep pixel clustering are presented in~\fref{fig:result_DIC}.

%% file: sections/fig_analysis_multiscale.tex
\begin{figure}[!t] 
\begin{center}
\begin{minipage}{0.19\linewidth}
\centerline{\includegraphics[scale=0.21]{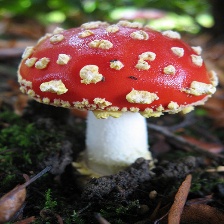}}
\end{minipage}
\begin{minipage}{0.19\linewidth}
\centerline{\includegraphics[scale=0.21]{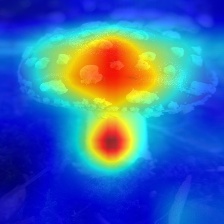}}
\end{minipage}
\begin{minipage}{0.19\linewidth}
\centerline{\includegraphics[scale=0.21]{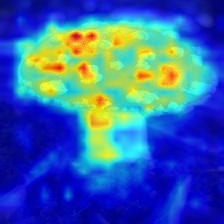}}
\end{minipage}
\begin{minipage}{0.19\linewidth}
\centerline{\includegraphics[scale=0.21]{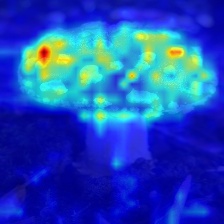}}
\end{minipage}
\begin{minipage}{0.19\linewidth}
\centerline{\includegraphics[scale=0.21]{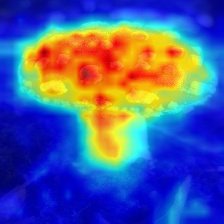}}
\end{minipage}
\\

\vspace{0.1cm}
\begin{minipage}{0.19\linewidth}
\centerline{\includegraphics[scale=0.21]{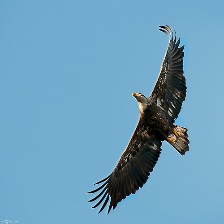}}
\end{minipage}
\begin{minipage}{0.19\linewidth}
\centerline{\includegraphics[scale=0.21]{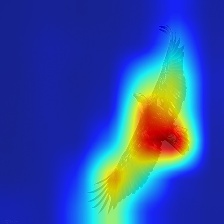}}
\end{minipage}
\begin{minipage}{0.19\linewidth}
\centerline{\includegraphics[scale=0.21]{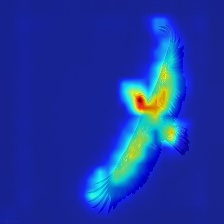}}
\end{minipage}
\begin{minipage}{0.19\linewidth}
\centerline{\includegraphics[scale=0.21]{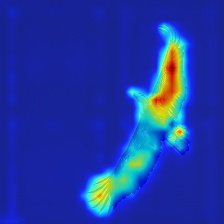}}
\end{minipage}
\begin{minipage}{0.19\linewidth}
\centerline{\includegraphics[scale=0.21]{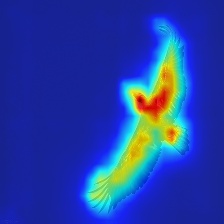}}
\end{minipage}
\\

\vspace{0.1cm}
\begin{minipage}{0.19\linewidth}
\centerline{(a)}
\end{minipage}
\begin{minipage}{0.19\linewidth}
\centerline{(b)}
\end{minipage}
\begin{minipage}{0.19\linewidth}
\centerline{(c)}
\end{minipage}
\begin{minipage}{0.19\linewidth}
\centerline{(d)}
\end{minipage}
\begin{minipage}{0.19\linewidth}
\centerline{(e)}
\end{minipage}
\caption{Visualization of class activation maps obtained from the multiscale object localization transformer using the ILSVRC-2012 dataset~\cite{ILSVRC15}. (a) Input image; (b) CAMs from the transformer employing the lowest resolution; (c) CAMs from the transformer using intermediate resolution; (d) CAMs from the transformer utilizing the highest resolution; (e) Combined class activation maps.}
\label{fig:analysis_multiscale}
\end{center}
\end{figure}

%% file: sections/4_results.tex
\section{EXPERIMENTS AND RESULTS}
\label{sec:results}

\subsection{Experimental Setting}
\label{sec:exp_setting}
\noindent \textbf{Dataset.}
We demonstrate the effectiveness of the proposed method using the publicly available ILSVRC-2012 dataset~\cite{ILSVRC15}. Following the experimental settings of previous works~\cite{Gao2021, cam2016}, the dataset, which contains 1,000 object classes, is divided into 1,281,167 images for training and 50,000 images for testing.

\input{sections/fig_result_ilsvrc}

\noindent \textbf{Evaluation metrics.}
We measure Top-1, Top-5, and GT-known localization accuracies to compare the performance of the proposed method with that of previous works, as outlined in the existing literature~\cite{Gao2021, cam2016}. While Top-1 and Top-5 localization accuracies evaluate classification and localization performances together, GT-known localization accuracy assesses only the localization performance.

Specifically, GT-known measures the proportion of images for which the computed intersection over union (IoU) scores are greater than 0.5. Top-1 and Top-5 localization accuracies are computed by measuring the proportion of images for which the Top-1 and Top-5 classification predictions are correct, respectively, in addition to having their IoU scores above 0.5.





\input{sections/tbl_result_ilsvrc_mobilenet}

\subsection{Results}
We present quantitative comparisons using the ILSVRC-2012 dataset~\cite{ILSVRC15} in~\tref{tab:result_ilsvrc}. Boldface and underline are used to denote the highest and second-highest scores in each metric, excluding our method without refinement. The proposed method is compared with convolutional neural network-based methods~\cite{cam2016, hideseek2017, Choe2019, Bae2020, Meng2021, spg2018, Xue2019, Zhang2020, i2c2020, Lu2020, Pan2021, Guo2021} as well as a vision transformer-based method~\cite{Gao2021}. The experimental results demonstrate that the proposed method outperforms previous works in Top-5 and GT-known localization metrics while achieving competitive accuracy in Top-1 localization. Specifically, the proposed method surpasses the previous state-of-the-art method by absolute 0.52\% and 0.59\% in Top-5 and GT-known localization accuracy, respectively, while achieving 0.51\% lower result in Top-1 localization. Please note that SLT utilizes distinct networks for classification and localization~\cite{Guo2021}, whereas MOLT employs a shared network for both tasks in order to enhance efficiency.

\fref{fig:result_ilsvrc} presents qualitative comparisons regarding localization using the ILSVRC-2012 dataset~\cite{ILSVRC15}. Red and green boxes are used to denote the ground-truth labels and the predicted object locations, respectively. The number in the right-top corner represents the computed IoU scores. Each row displays the input images, the results of TS-CAM~\cite{Gao2021}, the results of the proposed method without refinement, and the results of the proposed method. The figures demonstrate that the proposed method accurately localizes the corresponding objects while minimizing false positives and false negatives.

\begin{table}[!t]
\centering
\begin{minipage}{0.99\linewidth}
\caption{Ablation study of deep clustering-guided refinement using the ILSVRC-2012 dataset~\cite{ILSVRC15}.}
\label{tab:analysis_refinement}
\renewcommand{\arraystretch}{1.} 
\centering
\begin{tabular}{ >{\centering}m{0.38\textwidth}|  *{2}{>{\centering}m{0.11\textwidth}|} >{\centering\arraybackslash}m{0.14\textwidth} } 
\hline
Method &  Top-1 & Top-5 & GT-known \\ 
\hline\hline
MOLT w/o refinement & 54.85 & 65.47 & 68.71\\
MOLT   &   \textbf{55.19} & \textbf{65.92} & \textbf{69.21}\\
\hline  
\end{tabular}
\end{minipage}
\end{table}


\tref{tab:analysis_refinement} presents the ablation study of deep clustering-guided refinement using the ILSVRC-2012 dataset~\cite{ILSVRC15}. The refinement method improves Top-1, Top-5, and GT-known localization accuracies by 0.34\%, 0.45\%, and 0.50\%, respectively.


%% file: sections/fig_result_ilsvrc.tex
\begin{figure*}[!t] 
\begin{center}
\begin{minipage}{0.03\linewidth}
\centerline{(a)}
\end{minipage}
\begin{minipage}{0.11\linewidth}
\centerline{\includegraphics[scale=0.23]{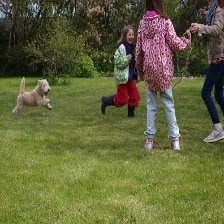}}
\end{minipage}
\begin{minipage}{0.11\linewidth}
\centerline{\includegraphics[scale=0.23]{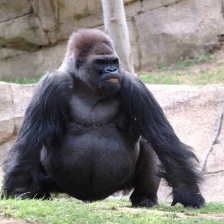}}
\end{minipage}
\begin{minipage}{0.11\linewidth}
\centerline{\includegraphics[scale=0.23]{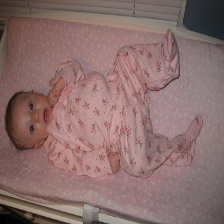}}
\end{minipage}
\begin{minipage}{0.11\linewidth}
\centerline{\includegraphics[scale=0.23]{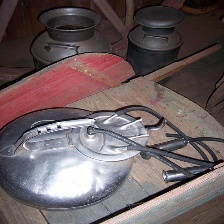}}
\end{minipage}
\begin{minipage}{0.11\linewidth}
\centerline{\includegraphics[scale=0.23]{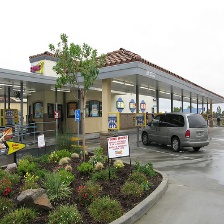}}
\end{minipage}
\begin{minipage}{0.11\linewidth}
\centerline{\includegraphics[scale=0.23]{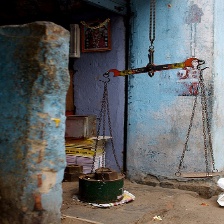}}
\end{minipage}
\begin{minipage}{0.11\linewidth}
\centerline{\includegraphics[scale=0.23]{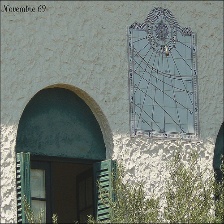}}
\end{minipage}
\\

\vspace{0.1cm}
\begin{minipage}{0.03\linewidth}
\centerline{(b)}
\end{minipage}
\begin{minipage}{0.11\linewidth}
\centerline{\includegraphics[scale=0.23]{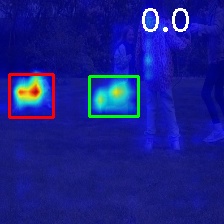}}
\end{minipage}
\begin{minipage}{0.11\linewidth}
\centerline{\includegraphics[scale=0.23]{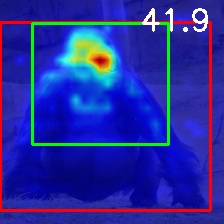}}
\end{minipage}
\begin{minipage}{0.11\linewidth}
\centerline{\includegraphics[scale=0.23]{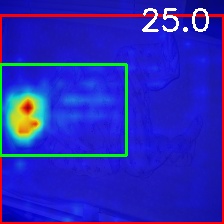}}
\end{minipage}
\begin{minipage}{0.11\linewidth}
\centerline{\includegraphics[scale=0.23]{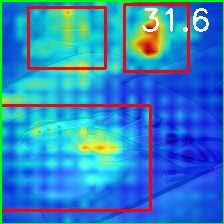}}
\end{minipage}
\begin{minipage}{0.11\linewidth}
\centerline{\includegraphics[scale=0.23]{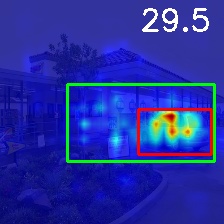}}
\end{minipage}
\begin{minipage}{0.11\linewidth}
\centerline{\includegraphics[scale=0.23]{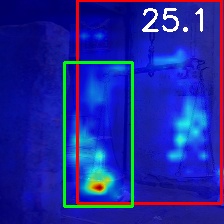}}
\end{minipage}
\begin{minipage}{0.11\linewidth}
\centerline{\includegraphics[scale=0.23]{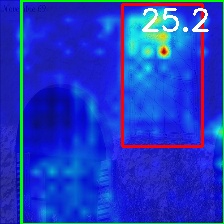}}
\end{minipage}
\\

\vspace{0.1cm}
\begin{minipage}{0.03\linewidth}
\centerline{(c)}
\end{minipage}
\begin{minipage}{0.11\linewidth}
\centerline{\includegraphics[scale=0.23]{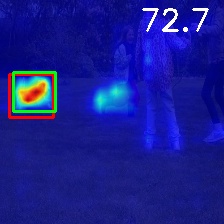}}
\end{minipage}
\begin{minipage}{0.11\linewidth}
\centerline{\includegraphics[scale=0.23]{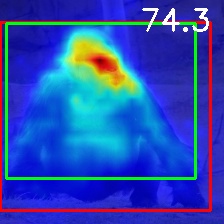}}
\end{minipage}
\begin{minipage}{0.11\linewidth}
\centerline{\includegraphics[scale=0.23]{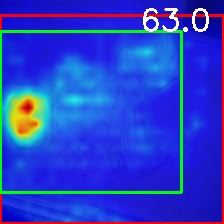}}
\end{minipage}
\begin{minipage}{0.11\linewidth}
\centerline{\includegraphics[scale=0.23]{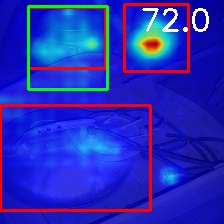}}
\end{minipage}
\begin{minipage}{0.11\linewidth}
\centerline{\includegraphics[scale=0.23]{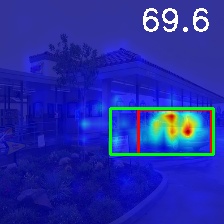}}
\end{minipage}
\begin{minipage}{0.11\linewidth}
\centerline{\includegraphics[scale=0.23]{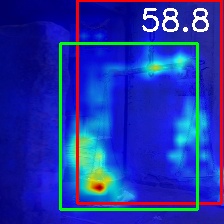}}
\end{minipage}
\begin{minipage}{0.11\linewidth}
\centerline{\includegraphics[scale=0.23]{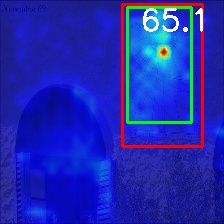}}
\end{minipage}
\\

\vspace{0.1cm}
\begin{minipage}{0.03\linewidth}
\centerline{(d)}
\end{minipage}
\begin{minipage}{0.11\linewidth}
\centerline{\includegraphics[scale=0.23]{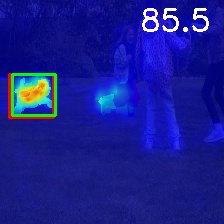}}
\end{minipage}
\begin{minipage}{0.11\linewidth}
\centerline{\includegraphics[scale=0.23]{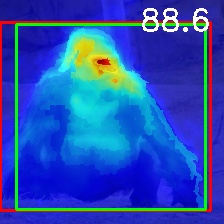}}
\end{minipage}
\begin{minipage}{0.11\linewidth}
\centerline{\includegraphics[scale=0.23]{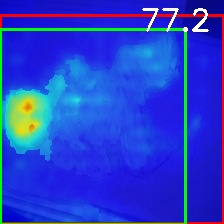}}
\end{minipage}
\begin{minipage}{0.11\linewidth}
\centerline{\includegraphics[scale=0.23]{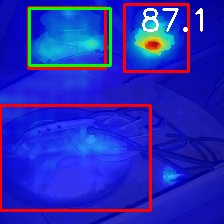}}
\end{minipage}
\begin{minipage}{0.11\linewidth}
\centerline{\includegraphics[scale=0.23]{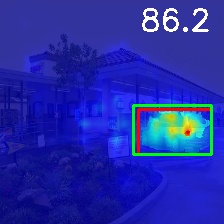}}
\end{minipage}
\begin{minipage}{0.11\linewidth}
\centerline{\includegraphics[scale=0.23]{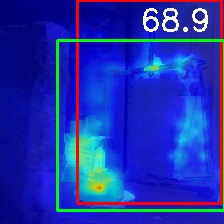}}
\end{minipage}
\begin{minipage}{0.11\linewidth}
\centerline{\includegraphics[scale=0.23]{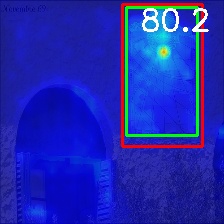}}
\end{minipage}
\caption{Qualitative comparison using the ILSVRC-2012 dataset~\cite{ILSVRC15}. (a) Image; (b) Result of TS-CAM~\cite{Gao2021}; (c) Result of the proposed method without refinement; (d) Result of the proposed method.}
\label{fig:result_ilsvrc}
\end{center}
\end{figure*}

%% file: sections/tbl_result_ilsvrc_mobilenet.tex
\begin{table}[!t]
\centering
\begin{minipage}{0.99\linewidth}
\caption{Quantitative comparison of Top-1, Top-5, and GT-known localization accuracies using the ILSVRC-2012 dataset~\cite{ILSVRC15}.}
\label{tab:result_ilsvrc}
\renewcommand{\arraystretch}{1.} 
\centering
\begin{tabular}{ >{\centering}m{0.2\textwidth}| >{\centering}m{0.18\textwidth}| *{2}{>{\centering}m{0.11\textwidth}|} >{\centering\arraybackslash}m{0.14\textwidth} } 
\hline
Method & Backbone & Top-1 & Top-5 & GT-known \\ 
\hline\hline
CAM~\cite{cam2016} &  & 43.35 & 54.44 & 58.97 \\
HaS~\cite{hideseek2017} & \multirow{2}{*}{MobileNetV1} & 42.73 & - & 60.12 \\
ADL~\cite{Choe2019} & \multirow{2}{*}{\cite{mobilenetv1}} & 43.01 & - & - \\
RCAM~\cite{Bae2020} &  & 44.78 & - & 61.69 \\
FAM~\cite{Meng2021} &  &  46.24 & - & 62.05 \\
\hline
CAM~\cite{cam2016}  &  & 46.29 & 58.19 & 62.68 \\
SPG~\cite{spg2018}  &  & 48.60 & 60.00 & 64.69 \\
DANet~\cite{Xue2019}  &  & 47.53 & 58.28 & - \\
PSOL~\cite{Zhang2020} &   & 54.82 & 63.25 & 65.21 \\
I2C~\cite{i2c2020} & InceptionV3 & 53.11 & 64.13 & 68.50 \\
GCNet~\cite{Lu2020} & \cite{inceptionv3} & 49.06 & 58.09 & - \\
RCAM~\cite{Bae2020} &  & 50.56 & - & 64.44 \\
SPA~\cite{Pan2021} &  & 52.73 & 64.27 & 68.33 \\
SLT~\cite{Guo2021} &  &   \textbf{55.70} & \underline{65.40} & 67.60 \\
FAM~\cite{Meng2021} &  &  \underline{55.24} & - & \underline{68.62}\\
\hline
TS-CAM~\cite{Gao2021} & DeiT-S  & {53.4} & {64.3} & {67.6}\\
MOLT (ours)  &  \cite{deit} & {55.19} & \textbf{65.92} & \textbf{69.21}\\
\hline  
\end{tabular}
\end{minipage}
\end{table}